\documentclass[conference]{IEEEtran}
\usepackage{cite}
\usepackage{amsmath,amssymb,amsfonts}
\usepackage{graphicx}
\usepackage{float}
\usepackage{textcomp}
\bstctlcite{IEEEexample:BSTcontrol}
\def\BibTeX{{\rm B\kern-.05em{\sc i\kern-.025em b}\kern-.08em
    T\kern-.1667em\lower.7ex\hbox{E}\kern-.125emX}}
\begin{document}

\title{
Operational AI Deployment Assurance:
Governance-State Orchestration Under
Threshold-Sensitive Deployment Conditions - A Governance Framework for High-Stakes AI Systems
}

\author{

\IEEEauthorblockN{
Khalid Adnan Alsayed
}

\IEEEauthorblockA{
Ducaltus | AI Assurance \& Governance\\
Newcastle upon Tyne, United Kingdom\\
School of Computing, Engineering \& Digital Technologies\\
Teesside University\\
Middlesbrough, United Kingdom\\
hello@ducaltus.com\\
s.khalid.adnan@gmail.com
}

}

\maketitle

\begin{abstract}

AI governance frameworks increasingly emphasize fairness, transparency, accountability, and lifecycle risk management in high-stakes domains. However, many current approaches remain observational, relying on static metric reporting, post-hoc auditing, and monitoring dashboards without directly governing deployment readiness, remediation progression, escalation states, or assurance-driven deployment control.
This paper introduces Operational AI Deployment Assurance (OADA), a governance framework for translating fairness disagreement, subgroup instability, threshold sensitivity, remediation outcomes, and operational uncertainty into deployment-oriented assurance decisions. Building on prior work on the Fairness Disagreement Index (FDI) and FairRisk-FDI, OADA reframes governance uncertainty as an operational concern within AI deployment pipelines rather than a byproduct of metric disagreement.
The framework introduces Deployment Assurance Scores, Deployment Readiness Classifications, Threshold Stability Zones, Governance Escalation States, and remediation-aware assurance progression. These constructs support lifecycle-oriented governance decisions across high-stakes settings by connecting evaluation outputs to deployment-state interpretation, reassessment, escalation, and operational control.
Through deployment-oriented evaluation across facial recognition systems, with discussion extended to healthcare AI as a representative high-stakes domain, the paper demonstrates how systems may appear acceptable under isolated fairness or performance metrics while still exhibiting instability that affects deployment readiness. The proposed framework positions operational deployment assurance as a governance layer between evaluation and real-world AI deployment.
\end{abstract}

\section{Introduction}
Artificial Intelligence systems are increasingly being deployed in high-stakes operational environments, including healthcare diagnostics, law enforcement, financial decision-making, critical infrastructure management, and public-sector governance \cite{b18,b19}. In these contexts, AI systems now participate in decisions capable of affecting human safety, legal outcomes, institutional accountability, and public trust. As deployment expands across socially sensitive and safety-critical domains, concerns surrounding fairness, transparency, accountability, robustness, and governance have become central challenges in the responsible deployment of machine learning systems \cite{b1,b28}.

Recent years have witnessed substantial advances in AI governance frameworks and responsible AI initiatives aimed at addressing these concerns \cite{b18}. Regulatory and governance efforts such as the NIST AI Risk Management Framework, the European Union AI Act, and emerging ISO/IEC governance standards have emphasized lifecycle risk management, trustworthy AI deployment, accountability mechanisms, and continuous monitoring for high-risk systems.

Despite these advances, many existing governance approaches remain primarily observational in nature \cite{b18,b25}. In practice, governance workflows frequently rely on static metric reporting, post-hoc auditing procedures, fairness dashboards, periodic compliance reviews, and aggregate performance summaries that observe model behaviour without governing deployment readiness or lifecycle assurance \cite{b35}. As a result, AI systems may satisfy isolated fairness or performance criteria while still exhibiting deployment instability, subgroup disparities, threshold sensitivity, remediation uncertainty, or governance ambiguity that undermine deployment trustworthiness \cite{b2,b33}.

This limitation becomes critical in high-stakes environments where deployment decisions cannot reasonably depend upon isolated evaluation metrics alone \cite{b16,b17}. Different fairness metrics frequently produce conflicting conclusions regarding the same model, especially when subgroup performance trade-offs emerge across race, gender, age, or intersectional demographic categories \cite{b5}. Prior work on the Fairness Disagreement Index (FDI) demonstrated that disagreement between fairness metrics represents a measurable property of machine learning evaluation rather than a purely interpretive concern \cite{b39}. Building upon this foundation, FairRisk-FDI further showed that disagreement amplification may propagate into deployment-oriented uncertainty, affecting governance confidence and deployment risk assessment in high-stakes systems \cite{b40}.

However, while existing literature extensively studies fairness evaluation, bias mitigation, explainability, and AI risk management, limited attention has been given to deployment assurance as an adaptive governance process \cite{b25}. Current governance approaches rarely operationalize deployment readiness assessment, remediation progression, governance escalation handling, assurance-state transitions, deployment gating, or remediation-aware reassessment within unified governance workflows \cite{b21,b36}.

This paper introduces Operational AI Deployment Assurance (OADA), a governance-oriented framework designed to operationally manage deployment readiness under evolving uncertainty conditions. Rather than treating governance as a static compliance mechanism, OADA conceptualizes deployment oversight as a continuously adaptive operational process connecting evaluation outputs, disagreement behaviour, remediation progression, governance escalation, and deployment-state interpretation throughout AI lifecycles.

The proposed framework introduces several operational governance constructs. Deployment Assurance Scores (DAS) represent deployment confidence under evolving operational conditions. Deployment Readiness Classifications (DRC) operationalize deployment-state interpretation under varying instability conditions. Threshold Stability Zones (TSZ) model governance sensitivity under threshold variation, while Governance Escalation States (GES) represent operational intervention pathways under elevated deployment uncertainty. In addition, remediation-aware governance progression enables deployment-state reassessment following mitigation and stabilization procedures.

The primary contributions of this paper are as follows:
\begin{itemize}
\item Introduction of Operational AI Deployment Assurance (OADA), a framework for governance-oriented deployment assurance under evolving uncertainty conditions.

\item Formalization of deployment-oriented governance constructs including Deployment Assurance Scores (DAS), Deployment Readiness Classifications (DRC), Threshold Stability Zones (TSZ), Governance Escalation States (GES), and remediation-aware deployment progression.

\item Operational distinction between monitoring and deployment assurance.

\item Experimental deployment-oriented evaluation across healthcare and facial recognition systems demonstrating instability-sensitive governance behaviour under varying threshold and remediation conditions.

\item Positioning of deployment assurance as a governance layer between evaluation and real-world AI deployment.

\end{itemize}
\section{Limitations of Observational AI Governance}
Contemporary AI governance frameworks have significantly advanced the discussion surrounding fairness, accountability, transparency, robustness, and responsible deployment in high-stakes machine learning systems \cite{b18,b21}. Regulatory initiatives, auditing methodologies, fairness toolkits, and model monitoring platforms have collectively contributed towards improved visibility into model behaviour and subgroup performance disparities. However, despite these advances, much of current AI governance remains fundamentally observational rather than operational in nature \cite{b22,b28}.

In many real-world deployment settings, governance processes primarily consist of metric reporting, dashboard visualisation, post-hoc auditing, periodic compliance assessment, and static evaluation summaries \cite{b22,b25}. While these mechanisms improve visibility into model behaviour, they frequently observe deployment conditions without governing deployment readiness, remediation progression, escalation handling, or deployment-state transitions \cite{b25}.

This limitation becomes increasingly significant as AI systems move into operationally sensitive domains where deployment decisions carry legal, financial, medical, or societal consequences \cite{b16,b19}. In such environments, acceptable aggregate accuracy or isolated fairness outcomes do not necessarily imply operational deployment readiness \cite{b39,b40}. Models may demonstrate strong overall performance while simultaneously exhibiting subgroup instability, threshold sensitivity, demographic trade-offs, or conflicting fairness evaluations that introduce uncertainty into deployment decisions.

A major limitation of current governance approaches lies in their evaluation-centric structure \cite{b5,b6}. Fairness evaluation commonly relies upon isolated metrics such as false positive rate disparity, false negative rate disparity, demographic parity, equalized odds, calibration measures, or subgroup accuracy comparison \cite{b3,b4}. However, prior research has shown that fairness metrics may produce conflicting conclusions regarding the same system under varying demographic compositions, threshold selections, operational conditions, or domain-specific priorities \cite{b5,b7}. Consequently, governance decisions derived from isolated evaluation metrics may inherit unresolved ambiguity regarding whether a system should remain deployable, restricted, reassessed, or operationally unsafe.

Existing governance frameworks also rarely operationalize the relationship between evaluation outcomes and deployment control decisions \cite{b18,b25}. In many deployment pipelines, governance terminates at reporting rather than progressing into structured governance mechanisms such as deployment gating, reassessment procedures, remediation escalation, threshold stabilization, or governance-state transitions \cite{b25}. Even when deployment risks or subgroup disparities are identified, governance frameworks frequently provide limited operational guidance regarding how remediation progression should be monitored, when escalation should occur, or how deployment assurance should evolve following mitigation interventions.

This distinction reflects a broader separation between monitoring and assurance within current AI governance ecosystems \cite{b22,b25,b26}. Monitoring systems primarily observe and report model behaviour during or after deployment, whereas assurance-oriented governance requires mechanisms capable of governing deployment trustworthiness itself \cite{b25}. Monitoring identifies system characteristics; assurance governs whether deployment conditions remain operationally acceptable under evolving uncertainty, subgroup instability, remediation outcomes, and governance constraints.

These limitations suggest that existing governance paradigms remain structurally incomplete for high-stakes operational AI deployment. As AI systems increasingly participate in socially consequential environments, governance mechanisms must evolve beyond observational reporting toward operational deployment assurance frameworks capable of integrating evaluation uncertainty, remediation progression, lifecycle governance, deployment readiness assessment, and assurance-driven deployment control into adaptive governance processes \cite{b25,b40}.
\section{Related Work}
\subsection{Algorithmic Fairness and Subgroup Evaluation}

Research on algorithmic fairness has expanded substantially in recent years as machine learning systems have become increasingly integrated into socially consequential and high-stakes environments. Early studies demonstrated that machine learning systems may exhibit unequal performance across demographic groups, particularly with respect to race, gender, age, and intersectional subgroup membership \cite{b2,b5}. These disparities have been observed across multiple application domains, including facial recognition, healthcare, criminal justice, hiring systems, financial decision-making, and automated content moderation, raising concerns regarding discrimination, accountability, and unequal operational harm \cite{b11,b13,b16}.

A substantial body of literature has focused on defining and evaluating statistical notions of fairness in machine learning systems. Common fairness metrics include demographic parity, equal opportunity, equalized odds, predictive parity, calibration consistency, false positive rate disparity, and false negative rate disparity \cite{b3,b4}. However, prior work has demonstrated that many fairness definitions are mathematically incompatible under differing base rates and operational conditions, leading to trade-offs between competing fairness objectives \cite{b5,b6}. Consequently, determining whether a model should be considered fair frequently depends upon the selected fairness metric, threshold configuration, subgroup distribution, and deployment context.

Research has also increasingly emphasized the importance of subgroup-aware and intersectional evaluation. Aggregate performance metrics may conceal substantial disparities affecting minority or intersectional populations, particularly in datasets characterized by demographic imbalance or uneven representation \cite{b2,b9}. Intersectional fairness research has further demonstrated that evaluating demographic categories independently may fail to capture compounded disparities emerging across combined subgroup identities such as race and gender simultaneously \cite{b2,b29}. These findings have motivated broader interest in subgroup reliability, fairness auditing, threshold analysis, and deployment-aware evaluation strategies for high-stakes AI systems.

More recent literature has increasingly questioned whether isolated fairness metrics alone provide sufficient foundations for deployment-oriented governance decisions. Studies examining fairness trade-offs, uncertainty, and operational reliability have argued that fairness evaluation should be contextualized within broader governance, risk management, and deployment frameworks rather than treated as an isolated statistical exercise \cite{b28,b39,b40}. This shift reflects growing recognition that fairness evaluation represents only one component within a larger ecosystem of trustworthy and operationally governed AI deployment.

\subsection{AI Governance and Trustworthy AI}

Alongside advances in algorithmic fairness research, significant attention has been directed toward the development of AI governance frameworks intended to support trustworthy, accountable, and responsible deployment of AI systems. Governments, standards organizations, regulatory bodies, and research institutions have increasingly recognized that high-stakes AI systems require governance mechanisms extending beyond conventional performance evaluation due to the societal, legal, and operational risks associated with automated decision-making \cite{b18,b28}.

A wide range of governance initiatives have emerged in response to these concerns. Regulatory and policy-oriented frameworks such as the European Union AI Act, the OECD AI Principles, UNESCO’s Recommendation on the Ethics of Artificial Intelligence, and the NIST Artificial Intelligence Risk Management Framework (AI RMF) emphasize principles including transparency, accountability, robustness, human oversight, fairness, safety, explainability, and continuous risk management across AI system lifecycles \cite{b18,b19,b20,b37}. Similarly, emerging standards efforts including ISO/IEC 42001 and related AI governance standards increasingly frame AI governance as a lifecycle-oriented process requiring ongoing monitoring, documentation, risk assessment, and organizational accountability \cite{b21,b27}.

Within the research community, trustworthy AI has become a major interdisciplinary area encompassing fairness, explainability, robustness, interpretability, privacy preservation, accountability mechanisms, and sociotechnical governance \cite{b28,b29}. Prior studies have argued that trustworthy AI deployment requires governance structures capable of addressing not only technical system behaviour but also organizational processes, human oversight responsibilities, deployment context, institutional accountability, and downstream societal impact \cite{b30,b31}. This broader governance perspective has increased interest in AI auditing methodologies, algorithmic accountability frameworks, documentation standards, impact assessments, and governance reporting systems \cite{b34,b36}.

However, despite the growing maturity of trustworthy AI and governance ecosystems, existing governance approaches remain heavily oriented toward documentation, reporting, monitoring, and compliance-centered oversight rather than operational deployment assurance. Many governance frameworks successfully identify governance objectives and high-level organizational principles while providing limited mechanisms for governing deployment-state transitions, remediation escalation, deployment gating, assurance progression, or operational deployment control under uncertainty \cite{b25,b27}.

These limitations suggest that while existing AI governance frameworks have substantially advanced responsible AI discourse, further work is required to operationalize governance as an adaptive deployment assurance process. There remains a need for governance-oriented frameworks capable of integrating fairness disagreement, remediation workflows, deployment readiness assessment, lifecycle governance progression, and operational deployment control into unified operational assurance infrastructures for high-stakes AI systems.

\subsection{MLOps, Monitoring, and Lifecycle Governance}

As machine learning systems have transitioned from experimental research settings into large-scale operational deployment environments, substantial attention has been directed toward Machine Learning Operations (MLOps), lifecycle management, and post-deployment monitoring infrastructures. MLOps frameworks aim to support reproducibility, scalability, deployment automation, model versioning, continuous integration and delivery, monitoring, and operational maintenance throughout machine learning system lifecycles \cite{b22,b23}. These developments have become increasingly important as organizations seek to operationalize AI systems within dynamic real-world environments characterized by evolving data distributions, infrastructure dependencies, regulatory constraints, and operational risk.

A major focus within the MLOps literature involves post-deployment monitoring and operational observability. Existing research has explored mechanisms for detecting concept drift, data drift, model degradation, calibration instability, performance decay, and anomalous system behaviour during deployment \cite{b24,b32}. Monitoring platforms frequently track prediction confidence, latency, subgroup performance variation, distributional shifts, and deployment reliability indicators to maintain visibility into deployed system behaviour over time \cite{b23,b36}. These monitoring capabilities represent important advances compared to static pre-deployment evaluation pipelines and have significantly improved operational awareness within deployed AI ecosystems.

Despite these advances, much of the existing MLOps and monitoring literature remains primarily focused on system observability rather than operational deployment assurance. Monitoring infrastructures are typically designed to identify and report operational conditions, but comparatively limited attention has been directed toward how governance systems should operationally respond once deployment instability, fairness disagreement, subgroup degradation, or threshold sensitivity are detected. In many existing deployment ecosystems, monitoring outputs remain informational rather than governance-executable, requiring manual interpretation without structured operational assurance mechanisms capable of governing deployment-state transitions or remediation progression.

This distinction highlights an important conceptual separation between monitoring infrastructures and assurance-oriented governance systems. Monitoring systems primarily provide visibility into operational conditions, whereas assurance-oriented governance requires mechanisms capable of determining whether deployment conditions remain operationally acceptable under evolving uncertainty, subgroup instability, and governance constraints. Existing monitoring pipelines may successfully detect deployment anomalies while lacking structured governance processes for escalation handling, conditional deployment restriction, remediation reassessment, or deployment-readiness reclassification.

These limitations suggest that although MLOps and monitoring ecosystems have substantially advanced operational visibility and deployment lifecycle management, additional governance-oriented assurance mechanisms are required to operationalize deployment control, remediation governance, lifecycle escalation handling, and assurance-state progression for high-stakes AI systems. This gap motivates the need for operational deployment assurance frameworks capable of integrating monitoring outputs, fairness disagreement, deployment instability, and governance decision-making into continuously governed operational deployment infrastructures.

\subsection{Fairness Disagreement and Operational Governance}

Recent research has increasingly questioned the assumption that isolated fairness metrics provide stable or universally reliable foundations for governance and deployment decision-making in machine learning systems. While traditional fairness evaluation approaches frequently treat individual fairness metrics as sufficient indicators of model fairness, multiple studies have demonstrated that differing fairness definitions may produce conflicting conclusions regarding the same system depending on subgroup composition, threshold selection, operational context, and underlying class distributions \cite{b5,b6,b12}. These findings have contributed toward broader recognition that fairness evaluation itself may represent an uncertain and context-dependent process rather than a singular or universally interpretable measurement task.

Building upon this broader literature, recent work on the Fairness Disagreement Index (FDI) introduced disagreement between fairness metrics as a measurable property of machine learning evaluation rather than a purely interpretive concern \cite{b39}. The FDI framework demonstrated that fairness metrics may diverge substantially in their assessment of the same system, particularly under varying threshold conditions and subgroup distributions, exposing instability within fairness evaluation processes themselves. This work argued that fairness disagreement should be treated as an operationally relevant property of model evaluation rather than an incidental byproduct of metric diversity.

Subsequent work on FairRisk-FDI extended this perspective by linking fairness disagreement to deployment-oriented governance risk \cite{b40}. Rather than treating fairness disagreement solely as an evaluation inconsistency, FairRisk-FDI demonstrated how disagreement between fairness metrics may propagate into deployment uncertainty and governance instability within high-stakes AI systems. The framework introduced deployment-oriented perspectives on fairness disagreement by incorporating decision-aware risk analysis, domain-sensitive deployment interpretation, and governance-oriented deployment classification concepts across healthcare and facial recognition domains.

These developments contribute toward an emerging shift from fairness evaluation toward deployment-oriented governance reasoning. Rather than asking solely whether a system satisfies isolated fairness criteria, recent work increasingly emphasizes broader questions concerning deployment trustworthiness, operational reliability, subgroup deployment risk, governance uncertainty, and lifecycle oversight under evolving deployment conditions \cite{b28,b40}. This shift aligns with growing recognition that high-stakes AI governance requires more than isolated fairness auditing and instead demands operational mechanisms capable of governing deployment decisions under uncertainty.

However, despite these advances, comparatively limited work has explored how fairness disagreement, deployment uncertainty, remediation progression, governance escalation, and operational lifecycle control may be integrated into unified operational assurance infrastructures. Existing fairness disagreement literature primarily focuses on evaluation instability and deployment risk identification rather than operational deployment governance mechanisms capable of continuously managing deployment readiness, remediation-state progression, assurance escalation, and deployment control throughout AI lifecycles.

\section{Operational AI Deployment Assurance (OADA)}
Existing AI governance approaches predominantly evaluate whether machine learning systems satisfy predefined fairness, performance, robustness, or compliance criteria at specific points within deployment lifecycles. While such approaches provide valuable visibility into model behaviour and deployment risk, they frequently remain observational in nature, emphasizing measurement, reporting, and post-hoc analysis rather than deployment governance.

This paper proposes Operational AI Deployment Assurance (OADA) as a governance-oriented framework designed to transition AI oversight from observational evaluation toward continuously governed deployment assurance. OADA conceptualizes AI governance not as a static compliance exercise, but as an active lifecycle process in which deployment trustworthiness is continuously assessed, reassessed, governed, and regulated under evolving uncertainty conditions.

\begin{figure}[t]
\centering
\includegraphics[width=\linewidth]{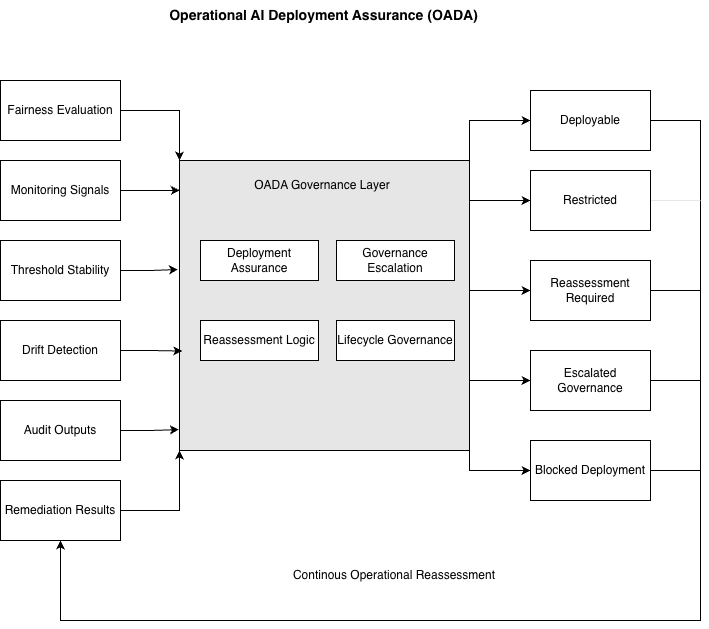}
\caption{Operational AI Deployment Assurance (OADA) lifecycle architecture integrating governance signals, deployment-state transitions, remediation-aware reassessment, and continuous operational oversight within high-stakes AI deployment environments.}
\label{fig:oada_architecture}
\end{figure}

Figure~\ref{fig:oada_architecture} illustrates the proposed Operational AI Deployment Assurance (OADA) lifecycle architecture. The framework integrates fairness evaluation, monitoring outputs, threshold stability assessment, remediation progression, and governance escalation into an adaptive deployment assurance process.

Table~\ref{tab:oada_constructs} contrasts monitoring-oriented governance with assurance-oriented governance within the proposed Operational AI Deployment Assurance (OADA) framework.

\begin{table}[t]
\caption{Operational Governance Constructs within OADA}
\label{tab:oada_constructs}
\centering
\begin{tabular}{|p{2.6cm}|p{2.0cm}|p{2.5cm}|}
\hline
\textbf{Construct} & \textbf{Purpose} & \textbf{Governance Role} \\
\hline
Deployment Assurance Score (DAS) & Represents deployment confidence under evolving conditions & Supports deployment readiness assessment \\
\hline
Threshold Stability Zones (TSZ) & Models governance sensitivity under threshold variation & Identifies instability-sensitive deployment regions \\
\hline
Governance Escalation & Represents escalation pathways under elevated uncertainty & Supports adaptive governance intervention \\
\hline
Remediation Progression & Tracks mitigation effectiveness and reassessment outcomes & Enables recovery-oriented governance \\
\hline
Deployment Readiness Classification & Represents deployment-state interpretation & Governs deployment-state transitions \\
\hline
\end{tabular}
\end{table}

Within OADA, assurance extends beyond observational analytics by governing deployment readiness, reassessment progression, escalation behaviour, and adaptive deployment-state control throughout AI lifecycles.

A central distinction underlying OADA is the separation between monitoring and assurance. Monitoring systems primarily observe operational conditions by reporting model behaviour, subgroup performance variation, drift indicators, or deployment metrics \cite{b23,b25}. Assurance-oriented governance, by contrast, governs whether operational deployment conditions remain sufficiently trustworthy to justify continued deployment under evolving uncertainty. While monitoring identifies deployment characteristics, assurance determines whether deployment should proceed, remain restricted, undergo remediation, enter reassessment, or transition into escalated governance states.

Assurance states may evolve as deployment conditions change, mitigation procedures are introduced, subgroup instability emerges, or governance uncertainty increases during deployment lifecycles.

Within OADA, governance transitions are triggered by evolving deployment conditions including fairness disagreement, subgroup instability, threshold sensitivity, operational drift, and unresolved remediation outcomes \cite{b39,b40}. These conditions may initiate reassessment, deployment restriction, escalation procedures, or deployment gating actions depending on operational severity.

OADA integrates existing fairness evaluation, monitoring, auditing, and governance mechanisms within a unified deployment assurance architecture. Fairness metrics, monitoring outputs, and audit results remain essential governance inputs, but are operationally interpreted as signals contributing toward adaptive deployment assurance decisions rather than isolated governance endpoints.

Table~\ref{tab:monitoring_vs_assurance} summarizes the principal governance distinctions introduced within the proposed Operational AI Deployment Assurance (OADA) framework.

\begin{table}[t]
\caption{Monitoring Versus Assurance within OADA}
\label{tab:monitoring_vs_assurance}
\centering
\begin{tabular}{|p{2.4cm}|p{2.0cm}|p{3.0cm}|}
\hline
\textbf{Dimension} & \textbf{Monitoring-Oriented Governance} & \textbf{Assurance-Oriented Governance} \\
\hline
Primary Function & Observe and report deployment behaviour & Govern deployment readiness and operational trustworthiness \\
\hline
Role in Deployment & Tracks operational conditions & Adapts governance intervention and deployment-state control \\
\hline
Response Capability & Passive observation and analytics & Active reassessment, escalation, restriction, and deployment control \\
\hline
Temporal Orientation & Primarily retrospective or observational & Adaptive and lifecycle-oriented \\
\hline
Governance Interpretation & Evaluation-focused oversight & Operational deployment assurance \\
\hline
Relationship to Deployment & Supports monitoring visibility & Governs deployment-state evolution \\
\hline
\end{tabular}
\end{table}

\subsection{Comparative Positioning}

Existing governance frameworks provide important foundations for trustworthy AI, monitoring, auditing, lifecycle oversight, and risk management. However, many existing approaches remain primarily observational and do not explicitly govern deployment-state transitions, escalation-sensitive governance handling, remediation progression, or threshold-sensitive assurance behaviour under evolving conditions.

Table~\ref{tab:comparative_positioning} compares OADA against representative governance and monitoring approaches across operational deployment assurance dimensions.

\begin{table}[t]
\caption{Comparative Positioning of OADA Against Existing Governance Approaches}
\label{tab:comparative_positioning}
\centering
\renewcommand{\arraystretch}{1.15}
\begin{tabular}{|p{2.4cm}|p{2.5cm}|p{2.5cm}|}
\hline
\textbf{Dimension} & \textbf{Existing Governance Frameworks} & \textbf{OADA} \\
\hline
Monitoring Support & Yes & Yes \\
\hline
Deployment-state Transition & Limited deployment formalisation & Explicit deployment states \\
\hline
Escalation Progression & High-level governance guidance & Structured escalation states \\
\hline
Remediation Progression & Partially addressed & Reassessment-oriented governance \\
\hline
Threshold Instability Modelling & Rarely operationalised & Explicit TSZ modelling \\
\hline
Deployment Assurance Aggregation & Not explicitly formalised & DAS-based governance interpretation \\
\hline
Operational Deployment Control & Primarily observational & Governance-state orchestration \\
\hline
\end{tabular}
\end{table}

While existing governance frameworks provide essential monitoring, auditing, lifecycle management, and risk governance capabilities, they do not explicitly govern deployment assurance as an adaptive governance-state process influenced by disagreement amplification, threshold instability, remediation progression, and escalation-sensitive operational behaviour. OADA extends these governance perspectives by introducing operational deployment-state interpretation and reassessment-oriented governance orchestration under instability-sensitive conditions.

\section{Operational Governance Constructs for Deployment Assurance}
Operational AI Deployment Assurance requires governance mechanisms capable of representing deployment trustworthiness, remediation progression, operational instability, and evolving deployment uncertainty within deployment environments. While existing governance approaches frequently rely upon isolated evaluation metrics or post-hoc reporting processes, the proposed framework introduces a set of operational governance constructs designed to support deployment-oriented assurance interpretation and lifecycle governance decision-making for high-stakes AI systems.

\begin{figure}[H]
\centering
\includegraphics[width=0.92\linewidth]{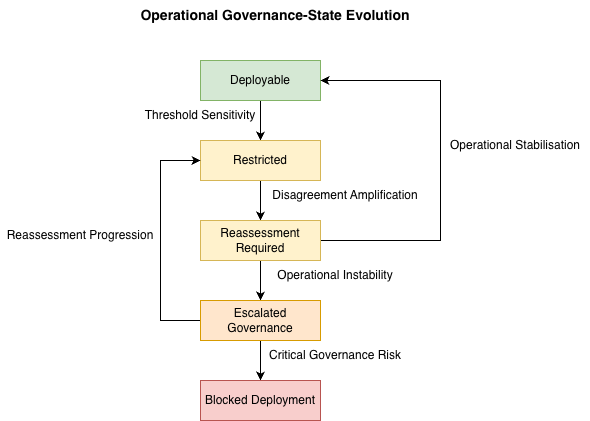}
\caption{Governance-State Transition Model within OADA}
\label{fig:governance_transition}
\end{figure}

Figure~\ref{fig:governance_transition} illustrates governance-state transitions within OADA. Deployment conditions evolve dynamically in response to threshold instability, disagreement amplification, remediation outcomes, operational uncertainty, and governance reassessment processes.

Table~\ref{tab:drc} defines the proposed deployment readiness classifications used within OADA to represent evolving governance conditions under high-stakes deployment environments.

\begin{table}[H]
\caption{Deployment Readiness Classification}
\label{tab:drc}
\centering
\scriptsize
\begin{tabular}{|p{2cm}|p{2.5cm}|p{2.5cm}|}
\hline
\textbf{Classification} & \textbf{Interpretation} & \textbf{Governance Action} \\
\hline
Deployable & Conditions remain sufficiently stable for deployment continuation & Continue deployment under standard governance monitoring \\
\hline
Restricted & Deployment conditions exhibit moderate instability or elevated sensitivity & Apply controlled deployment restrictions and increased oversight \\
\hline
Reassessment Required & Governance uncertainty requires reassessment before deployment continuation & Initiate reassessment and mitigation review procedures \\
\hline
Escalated Governance & Significant instability or disagreement amplification requires elevated governance intervention & Apply governance escalation and deployment restriction procedures \\
\hline
Blocked Deployment & Operational conditions are considered unsafe or insufficiently governed for deployment continuation & Prevent or suspend deployment until stabilization is achieved \\
\hline
\end{tabular}
\end{table}

The following subsections introduce the core governance constructs underlying the proposed Operational AI Deployment Assurance (OADA) framework, including Deployment Assurance Scores (DAS), Deployment Readiness Classifications (DRC), Threshold Stability Zones (TSZ), Governance Escalation States (GES), and remediation-aware governance progression mechanisms.

\subsection{Deployment Assurance Score (DAS)}

To support this broader governance perspective, the proposed framework introduces the Deployment Assurance Score (DAS), an operational governance construct intended to represent overall deployment assurance confidence under evolving deployment conditions.

Conceptually, DAS reflects the degree of operational confidence associated with continued deployment of a machine learning system under current governance conditions. DAS is not intended to replace existing evaluation metrics, but rather to operationalize their collective governance implications within a unified assurance-oriented deployment framework.

The Deployment Assurance Score (DAS) is operationalized as a weighted governance aggregation construct:

\begin{equation}
DAS=\alpha(1-FDI)+\beta(1-\Delta FPR)+\gamma(1-\Delta FNR)+\delta(1-TSZ)
\end{equation}

where FDI represents fairness disagreement instability, $\Delta FPR$ and $\Delta FNR$ represent subgroup disparity magnitudes, TSZ represents threshold stability behaviour, and $\alpha$, $\beta$, $\gamma$, and $\delta$ represent governance-sensitive weighting coefficients satisfying:

\begin{equation}
\alpha + \beta + \gamma + \delta = 1
\end{equation}

Within the current experimental evaluation, weighting coefficients are operationally assigned using governance-oriented domain interpretation rather than optimisation-based calibration. Sensitivity analysis of adaptive weighting strategies remains part of future work.

DAS operates as part of an adaptive governance lifecycle rather than as a static evaluation outcome.

Crucially, DAS is intended to function as an operational governance construct rather than a standalone optimization target. Consequently, assurance interpretation within OADA remains dependent upon broader governance context, deployment conditions, remediation status, operational constraints, and domain-specific risk considerations.

\subsection{Deployment Readiness Classification (DRC)}

To support this requirement, the proposed framework introduces Deployment Readiness Classification (DRC), a governance-oriented construct designed to represent deployment-state conditions within adaptive deployment environments.

Within the proposed framework, deployment readiness classifications function as operational governance states derived from evolving deployment assurance conditions rather than static certification outcomes.

Importantly, DRC is intended to support governance interpretation rather than deterministic deployment automation.

Within OADA, deployment readiness classifications may include governance-oriented operational states such as Deployable, Conditional Deployment, Restricted Deployment, Remediation Required, Escalated Review, or Blocked Deployment.

This perspective allows governance systems to operationalize deployment progression under uncertainty. Systems may transition into remediation-required states following the emergence of subgroup instability or threshold degradation, subsequently progressing toward conditional or deployable states following successful remediation and reassessment procedures. Conversely, unresolved mitigation failures, disagreement amplification, or increasing operational instability may trigger escalation into restricted or blocked deployment conditions. Governance therefore becomes adaptive and progression-oriented rather than statically evaluative.

\subsection{Threshold Stability Zones (TSZ)}

To address this limitation, the proposed framework introduces Threshold Stability Zones (TSZ), an operational governance construct designed to represent threshold-sensitive deployment stability conditions within deployment environments.

Conceptually, TSZ represents regions of threshold behaviour associated with differing levels of deployment stability and governance sensitivity.

Threshold Stability Zones (TSZ) are approximated using the rate of disagreement variation across deployment thresholds:

\begin{equation}
TSZ=\left|\frac{\partial FDI}{\partial t}\right|
\end{equation}

where $t$ represents deployment threshold configuration and FDI represents fairness disagreement instability. Higher TSZ values indicate increased operational sensitivity under small threshold adjustments.

\begin{figure}[H]
\centering
\includegraphics[width=\linewidth]{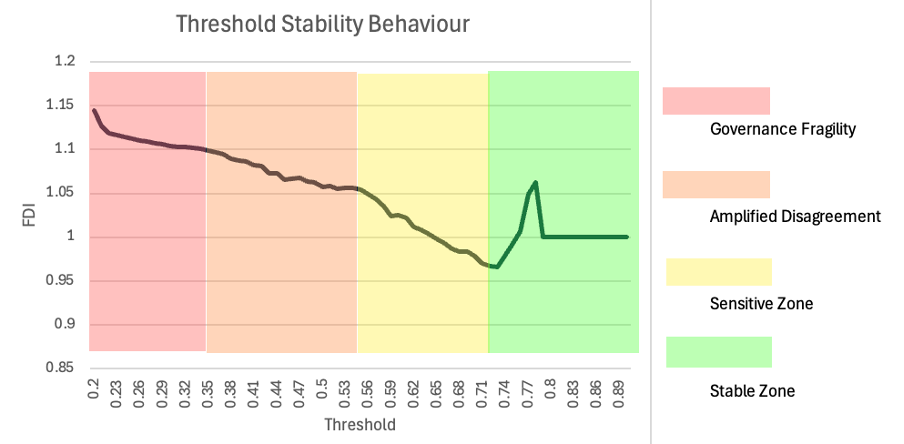}
\caption{Threshold Stability Zones (TSZ) illustrating governance sensitivity, disagreement amplification, and operational fragility under varying deployment threshold configurations.}
\label{fig:tsz}
\end{figure}

Table~\ref{tab:tsz} summarizes the proposed operational interpretation of Threshold Stability Zones (TSZ) under varying deployment threshold conditions within OADA.

\begin{table}[H]
\caption{Threshold Instability Interpretation within OADA}
\label{tab:tsz}
\centering
\scriptsize
\begin{tabular}{|p{2.2cm}|p{2.5cm}|p{2.5cm}|}
\hline
\textbf{Threshold Region} & \textbf{Operational Behaviour} & \textbf{Governance Interpretation} \\
\hline
Stable Zone & Minimal disagreement variation under threshold adjustment & Conditions remain sufficiently stable for deployment continuation \\
\hline
Sensitive Zone & Moderate threshold sensitivity with emerging disagreement variation & Increased governance monitoring and reassessment may be required \\
\hline
Amplified Disagreement Zone & Small threshold changes produce disproportionate disagreement escalation & Deployment conditions exhibit elevated instability and governance uncertainty \\
\hline
Governance Fragility Zone & Severe instability and disagreement amplification under threshold variation & Deployment conditions may require restriction, escalation, or suspension \\
\hline
\end{tabular}
\end{table}

Importantly, TSZ is not intended solely as a mathematical threshold-optimization mechanism. Rather, the construct functions as a governance-oriented operational abstraction designed to support deployment interpretation under uncertainty.

Within OADA, threshold stability assessment may influence broader deployment assurance interpretation, governance escalation severity, remediation requirements, and deployment-readiness classification.

\subsection{Governance Escalation States (GES)}

To address this limitation, the proposed framework introduces Governance Escalation States (GES), an operational governance construct designed to model escalation severity and governance response intensity within deployment environments.

Governance escalation behaviour within OADA is represented as a composite operational escalation function:

\begin{equation}
GES=f(FDI,\Delta FPR,\Delta FNR,TSZ,R_m)
\end{equation}

where FDI represents disagreement instability, $\Delta FPR$ and $\Delta FNR$ represent subgroup disparity magnitudes, TSZ represents threshold sensitivity behaviour, and $R_m$ represents remediation effectiveness under reassessment-oriented deployment conditions.

Within OADA, governance escalation states represent operational governance conditions triggered by evolving deployment assurance characteristics.

Within the proposed framework, escalation states may include governance-oriented conditions such as Monitored, Flagged, Restricted, Escalated, or Critical. Importantly, escalation progression within OADA is adaptive rather than static.

Within OADA, governance escalation states additionally interact with broader deployment assurance constructs including Deployment Assurance Scores, Deployment Readiness Classifications, Threshold Stability Zones, and remediation progression mechanisms.

Importantly, GES is not intended to automate governance decision-making independently of human oversight but to support interpretable lifecycle-oriented escalation handling.

Table~\ref{tab:ges} summarizes representative governance escalation conditions within OADA under varying operational instability and deployment uncertainty scenarios.

\begin{table}[H]
\caption{Governance Escalation Conditions}
\label{tab:ges}
\centering
\scriptsize
\begin{tabular}{|p{2.4cm}|p{2.5cm}|p{2.6cm}|}
\hline
\textbf{Escalation Level} & \textbf{Example Trigger} & \textbf{Operational Response} \\
\hline
Low & Minor subgroup instability or moderate threshold sensitivity & Continue monitoring and initiate targeted reassessment \\
\hline
Moderate & Persistent disagreement amplification or emerging operational instability & Apply deployment restrictions and increase governance oversight \\
\hline
High & Severe threshold instability, remediation failure, or escalating uncertainty & Escalate governance intervention and limit deployment conditions \\
\hline
Critical & Operational fragility or unacceptable deployment task conditions & Block or suspend deployment pending stabilization and reassessment \\
\hline
\end{tabular}
\end{table}

\subsection{Remediation Progression and Lifecycle Governance}

A major limitation of existing AI governance approaches is their tendency to treat remediation as an isolated corrective action rather than an adaptive governance lifecycle process. In practice, high-stakes AI systems frequently undergo multiple stages of reassessment, mitigation, threshold adjustment, subgroup rebalancing, monitoring enhancement, policy modification, or deployment restriction following the identification of governance concerns.

To address this limitation, the proposed framework introduces remediation-aware lifecycle governance as a core component of Operational AI Deployment Assurance. Within OADA, remediation is conceptualized not as a singular post-hoc intervention, but rather as an adaptive operational progression involving reassessment, assurance recovery, governance adaptation, escalation handling, and deployment-state evolution under changing deployment conditions.

Remediation progression within OADA is represented as the change in deployment assurance behaviour following mitigation intervention:

\begin{equation}
R_p = DAS(t+1)-DAS(t)
\end{equation}

where $R_p$ represents remediation progression and $DAS(t+1)$ and $DAS(t)$ represent post-remediation and pre-remediation deployment assurance states respectively.

Within OADA, remediation progression therefore functions as a lifecycle-oriented governance process in which deployment assurance conditions evolve dynamically following mitigation interventions. Consequently, remediation outcomes are not interpreted solely through isolated post-mitigation evaluation metrics, but rather through continuously evolving deployment assurance conditions integrated into broader governance lifecycles.

Importantly, remediation-aware governance within OADA does not assume that all deployment instability can or should be fully resolved through technical mitigation alone. Certain deployment conditions may remain operationally unacceptable despite repeated remediation attempts, particularly within environments characterized by persistent subgroup instability, unresolved disagreement amplification, or unacceptable operational risk. Consequently, governance systems must retain the ability to maintain deployment restrictions, escalate governance severity, or block deployment continuation when remediation progression fails to sufficiently restore deployment assurance.

\section{Experimental Operational Assurance Evaluation}
\subsection{Experimental Objectives}

This section evaluates the proposed Operational AI Deployment Assurance (OADA) framework under deployment-oriented operational conditions. The objective of the experimental evaluation is not to optimize predictive performance alone, but to examine how disagreement amplification, threshold instability, subgroup variation, remediation progression, and governance uncertainty influence deployment assurance interpretation within adaptive deployment environments.

The experimental analysis focuses on three primary objectives:

\begin{itemize}
\item Evaluating threshold-sensitive deployment instability under varying operational conditions.
\item Examining how governance-state interpretation evolves under disagreement amplification and subgroup instability.
\item Assessing remediation-aware deployment progression following mitigation interventions.
\end{itemize}

The experimental evaluation primarily focuses on facial recognition deployment environments using threshold-sensitive and remediation-aware governance analysis derived from prior fairness disagreement evaluation \cite{b39,b40}. Healthcare deployment scenarios are incorporated as representative high-stakes operational governance contexts based on prior FairRisk-FDI deployment-risk evaluation \cite{b13,b40}.

\subsection{Evaluation Domains and Datasets}

The experimental evaluation primarily utilizes facial recognition deployment environments to examine disagreement amplification, subgroup instability, threshold sensitivity, and remediation-aware governance behaviour under operational deployment conditions. Experimental analysis builds upon prior fairness disagreement evaluation conducted across demographic subgroup-sensitive facial recognition systems under varying deployment thresholds \cite{b15,b39}.

To support broader deployment-oriented governance interpretation, healthcare machine learning environments are additionally referenced as representative high-stakes operational AI settings where deployment instability, subgroup reliability variation, and governance uncertainty may influence real-world decision-making processes \cite{b16,b17}.

\subsubsection{Experimental Configuration}

The experimental evaluation utilizes deployment-oriented fairness disagreement analysis derived from prior threshold-sensitive and remediation-aware subgroup evaluation across facial recognition systems \cite{b39,b40}. Table~\ref{tab:experimental_config} summarizes the primary experimental configurations used within the operational governance evaluation.

\begin{table}[H]
\caption{Experimental Configuration Summary}
\label{tab:experimental_config}
\centering
\scriptsize
\begin{tabular}{|p{4.0cm}|p{4.0cm}|}
\hline
\textbf{Component} & \textbf{Configuration} \\
\hline
Facial Recognition Dataset & FairFace \\
\hline
Facial Recognition Evaluation & Demographic subgroup fairness analysis \\
\hline
Baseline Model & MTLVIFR (Multi-Task Learning Variation-Invariant Face Recognition) \\
\hline
Remediation Method 1 & Balanced Batch Sampling \\
\hline
Remediation Method 2 & Focal Loss Mitigation \\
\hline
Threshold Evaluation Range & 0.20--0.90 \\
\hline
Governance Evaluation Focus & FDI instability, TSZ behaviour, DRC transitions \\
\hline
Healthcare Context & High-stakes deployment interpretation reference \\
\hline
Experimental Objective & Operational deployment assurance evaluation \\
\hline
\end{tabular}
\end{table}

\subsection{Threshold Stability and Governance Sensitivity Analysis}

\subsubsection{Threshold Stability Behaviour}

Threshold variation analysis was conducted to examine instability-sensitive deployment behaviour under evolving operational conditions \cite{b5,b12}. Experimental evaluation focused on how small threshold adjustments influenced disagreement amplification, subgroup stability, deployment interpretation, and governance sensitivity across deployment environments.

Threshold analysis refers to decision-threshold variation applied during subgroup-sensitive deployment evaluation under fairness disagreement conditions.

The analysis demonstrated that deployment interpretation may remain relatively stable under certain threshold regions while rapidly destabilizing within instability-sensitive operational zones. These findings support the proposed Threshold Stability Zone (TSZ) construct by illustrating how governance interpretation may evolve non-linearly under small deployment configuration changes.

\subsubsection{Deployment Readiness Transition Analysis}

Deployment-state interpretation was further evaluated under varying disagreement and threshold conditions using the proposed Deployment Readiness Classification (DRC) framework \cite{b39,b40}. Experimental analysis examined how systems transitioned between deployable, restricted, reassessment-required, escalated-governance, and blocked-deployment states under evolving operational instability conditions.

The results demonstrated that governance interpretation may evolve dynamically as disagreement amplification, subgroup instability, or threshold sensitivity increases. These findings support the proposed governance-state transition model by illustrating that deployment readiness may function as an adaptive operational state rather than a static evaluative outcome.

Table~\ref{tab:threshold_transition} summarizes deployment-state interpretation under varying disagreement-sensitive threshold conditions within OADA.

\begin{table}[H]
\caption{Deployment Readiness Transition Analysis under Threshold Variation}
\label{tab:threshold_transition}
\centering
\scriptsize
\begin{tabular}{|p{1.0cm}|p{1.5cm}|p{1.5cm}|p{1.5cm}|p{1.3cm}|}
\hline
\textbf{Threshold Range} & \textbf{FDI Behaviour} & \textbf{Stability Interpretation} & \textbf{Deployment Conditions} & \textbf{DRC State} \\
\hline
0.20--0.35 & High disagreement behaviour & Severe instability-sensitive behaviour & Elevated governance uncertainty & Blocked Deployment \\
\hline
0.35--0.55 & Amplified disagreement behaviour & Governance-sensitive instability & Escalation-sensitive deployment conditions & Escalated Governance \\
\hline
0.55--0.72 & Moderate disagreement & Threshold-sensitive deployment behaviour & Reassessment-oriented governance conditions & Reassessment Required \\
\hline
0.72--0.78 & Relatively stabilised disagreement behaviour & Controlled operational stability & Restricted deployment conditions & Restricted \\
\hline
0.78--0.90 & Stabilised disagreement behaviour & Operationally stable deployment conditions & Acceptable deployment stability & Deployable \\
\hline
\end{tabular}
\end{table}

The results demonstrate that small threshold adjustments may substantially alter deployment-state interpretation under disagreement-sensitive operational conditions. These findings support the proposed Deployment Readiness Classification (DRC) framework by illustrating that deployment readiness may evolve dynamically under varying instability conditions rather than functioning as a static evaluation outcome.

\subsubsection{Governance Escalation Behaviour}

Governance escalation analysis examined how instability amplification and unresolved deployment uncertainty influenced operational governance intervention pathways within OADA. Experimental observations demonstrated that elevated disagreement conditions frequently propagated into reassessment escalation, deployment restriction, or governance-sensitive operational states under unstable threshold regions.

These findings support the proposed Governance Escalation States (GES) construct by demonstrating that deployment governance may require adaptive intervention mechanisms capable of responding dynamically to evolving instability conditions rather than relying solely upon static compliance-oriented evaluation procedures.

\subsection{Remediation-Aware Deployment Reassessment}

\subsubsection{Baseline Deployment Conditions}

Baseline deployment analysis evaluated operational governance behaviour prior to mitigation intervention. Experimental evaluation examined subgroup instability, disagreement amplification, deployment sensitivity, and deployment readiness interpretation under unmitigated operational conditions.

The baseline analysis demonstrated that acceptable aggregate performance may coexist with instability-sensitive deployment behaviour, subgroup disparities, and elevated governance uncertainty under varying deployment conditions.

\subsubsection{Post-Mitigation Assurance Reclassification}

Mitigation-aware evaluation examined how deployment assurance behaviour evolved following remediation interventions. Experimental analysis compared baseline deployment conditions against mitigation-oriented deployment progression under subgroup-sensitive operational conditions.

The analysis demonstrated that mitigation outcomes may influence deployment assurance non-linearly. Certain interventions improved subgroup stability and reduced disagreement amplification, while other mitigation strategies produced limited assurance recovery despite isolated performance improvements.

\begin{table}[H]
\caption{Remediation-Aware Deployment Progression Analysis}
\label{tab:remediation_progression}
\centering
\scriptsize
\begin{tabular}{|p{1.5cm}|p{0.8cm}|p{0.6cm}|p{0.6cm}|p{0.6cm}|p{0.6cm}|p{1.2cm}|}
\hline
\textbf{Model} & \textbf{Accuracy} & \textbf{Mean FPR} & \textbf{Mean FNR} & \textbf{FPR Gap} & \textbf{FNR Gap} & \textbf{Governance Outcome} \\
\hline
Baseline MTLVIFR & 0.7812 & 0.1084 & 0.4288 & 0.3043 & 0.6944 & Escalated Governance \\
\hline
Balanced Batch Sampling & 0.7986 & 0.0734 & 0.4014 & 0.2063 & 0.4235 & Restricted \\
\hline
Focal Loss Mitigation & 0.7507 & 0.1173 & 0.5076 & 0.3043 & 0.7014 & Reassessment Required \\
\hline
\end{tabular}
\end{table}

The results demonstrate that mitigation progression does not necessarily produce linear assurance recovery under operational governance evaluation. While balanced batch sampling \cite{b10} improved subgroup stability and reduced disagreement-sensitive deployment instability, focal loss mitigation \cite{b14} produced comparatively weaker operational assurance recovery despite mitigation intervention. These findings support the proposed reassessment-oriented governance perspective within OADA by illustrating that deployment assurance must remain continuously reassessed following remediation progression rather than statically assumed following isolated mitigation improvements.

\subsubsection{Quantitative Governance Trace}

To further illustrate operational governance behaviour within OADA, Table~\ref{tab:governance_trace} presents a quantitative governance trace derived from the remediation-aware deployment evaluation. The trace integrates disagreement instability, subgroup disparity behaviour, threshold sensitivity, deployment assurance interpretation, and resulting governance-state outcomes across mitigation conditions.

\begin{table}[H]
\caption{Quantitative Governance Trace under Remediation-Aware Deployment Evaluation}
\label{tab:governance_trace}
\centering
\scriptsize
\begin{tabular}{|p{1.5cm}|p{0.5cm}|p{0.6cm}|p{0.6cm}|p{0.5cm}|p{0.5cm}|p{1.4cm}|}
\hline
\textbf{Model} & \textbf{FDI} & \textbf{$\Delta$FPR} & \textbf{$\Delta$FNR} & \textbf{TSZ} & \textbf{DAS} & \textbf{DRC} \\
\hline
Baseline (MTLVIFR) & 0.68 & 0.304 & 0.694 & 0.42 & 0.48 & Escalated Governance \\
\hline
Balanced Batch Sampling & 0.41 & 0.206 & 0.424 & 0.28 & 0.71 & Restricted \\
\hline
Focal Loss Mitigation & 0.62 & 0.304 & 0.701 & 0.39 & 0.52 & Reassessment Required \\
\hline
\end{tabular}
\end{table}

The governance trace demonstrates how operational deployment interpretation evolves under changing remediation conditions rather than remaining fixed under static evaluation outputs. Balanced Batch Sampling substantially reduced disagreement instability and threshold-sensitive operational behaviour compared to the baseline model, resulting in improved deployment assurance interpretation under OADA. Conversely, Focal Loss mitigation failed to meaningfully reduce disagreement amplification despite remediation intervention, leaving deployment assurance behaviour within reassessment-oriented governance conditions.

These findings illustrate how OADA operationalizes deployment assurance as a continuously evolving governance process influenced by disagreement instability, subgroup-sensitive operational behaviour, threshold sensitivity, and remediation progression rather than isolated predictive performance metrics alone.

\subsubsection{Counterintuitive Deployment Behaviour}

A counterintuitive finding emerged from the remediation-aware deployment analysis. Although Balanced Batch Sampling produced only moderate improvements in aggregate predictive performance, it substantially reduced disagreement amplification and threshold-sensitive instability compared to the baseline deployment condition. Conversely, Focal Loss mitigation failed to meaningfully improve deployment assurance behaviour despite mitigation intervention, leaving disagreement instability and subgroup-sensitive operational behaviour elevated under reassessment-oriented governance conditions.

These findings suggest that remediation strategies optimised primarily for predictive performance may not necessarily improve deployment assurance under instability-sensitive operational conditions. In contrast, mitigation approaches that reduce disagreement amplification and threshold-sensitive governance instability may produce stronger deployment assurance outcomes despite comparatively smaller gains in aggregate predictive performance.

\subsubsection{Deployment-State Recovery Following Remediation}

Post-mitigation reassessment examined how deployment-state interpretation evolved following remediation progression within OADA. Experimental analysis evaluated whether mitigation interventions sufficiently stabilized operational conditions to support deployment-state recovery under reassessment-oriented governance evaluation.

The results demonstrated that remediation progression may alter governance interpretation over time, supporting the proposed reassessment-oriented deployment assurance model within adaptive deployment environments.

\subsection{Operational Governance Interpretation}

Collectively, the experimental findings demonstrate that deployment assurance cannot be reliably inferred from isolated predictive or fairness evaluation metrics alone \cite{b38}. Across both deployment domains, governance interpretation evolved dynamically under changing threshold conditions, subgroup instability, disagreement amplification, and remediation progression.

The results support the broader OADA perspective that operational deployment assurance requires adaptive governance mechanisms capable of integrating instability-sensitive deployment interpretation, reassessment progression, governance escalation, and remediation-aware deployment control within deployment lifecycles.

Rather than functioning solely as observational governance outputs, fairness evaluation, monitoring signals, remediation outcomes, and operational instability conditions become governance-driving operational variables influencing deployment-state evolution over time.

\section{Discussion}

\subsection{Monitoring Versus Assurance}

Current AI governance ecosystems frequently emphasize monitoring, auditing, fairness reporting, and post-deployment observability mechanisms \cite{b22,b35,b36}. While these approaches improve visibility into model behaviour, they do not necessarily operationalize deployment readiness management under evolving instability-sensitive conditions. The experimental findings presented in this paper demonstrate that deployment interpretation may evolve dynamically under threshold variation, disagreement amplification, subgroup instability, and remediation progression even when isolated predictive or fairness metrics appear acceptable.

Within OADA, monitoring and assurance serve distinct governance roles. Monitoring mechanisms observe operational behaviour, whereas assurance-oriented governance manages deployment readiness, escalation handling, reassessment progression, and deployment-state transitions under evolving operational conditions. This distinction becomes increasingly important in high-stakes deployment environments where governance decisions may carry legal, medical, financial, or societal consequences.

\subsection{Deployment Assurance as Governance Infrastructure}

The proposed OADA framework positions deployment assurance as an operational governance infrastructure layer situated between model evaluation and real-world deployment. Rather than treating governance as a static compliance exercise, the framework operationalizes deployment readiness as a continuously evolving lifecycle process influenced by disagreement amplification, threshold instability, remediation outcomes, governance escalation, and reassessment-sensitive operational behaviour.

The experimental findings support the broader argument that deployment assurance cannot be reliably inferred from isolated fairness metrics, aggregate predictive performance, or static audit outputs alone. Across both threshold-sensitive and remediation-aware evaluation settings, deployment interpretation evolved dynamically under changing operational conditions. These findings suggest that operational governance systems must increasingly support lifecycle-oriented reassessment, instability-sensitive deployment interpretation, and adaptive deployment assurance workflows.

Importantly, the framework does not seek to replace existing fairness evaluation, auditing, monitoring, or trustworthy AI methodologies \cite{b18,b28}. Instead, OADA integrates these mechanisms within an operational governance structure capable of supporting deployment-state interpretation, escalation-sensitive governance handling, and remediation-aware assurance progression across deployment lifecycles.

\subsection{Limitations and Future Work}

Several limitations remain within the current study. The experimental evaluation focused primarily on facial recognition and healthcare machine learning environments, and broader validation across additional operational domains remains necessary. Furthermore, the current framework primarily evaluates deployment assurance under threshold-sensitive and remediation-aware operational conditions rather than fully real-time deployment infrastructures.

Future work may explore longitudinal deployment evaluation, real-time governance orchestration, automated escalation-sensitive deployment management, temporal instability modelling, and integration with enterprise-scale deployment assurance systems. Additional research may also investigate deployment assurance behaviour under continuously adaptive learning environments, foundation models, multimodal AI systems, and large-scale operational AI infrastructures.

\section{Conclusion}

This paper introduced Operational AI Deployment Assurance (OADA), a governance-oriented framework for continuously managing deployment readiness under disagreement-sensitive operational conditions. Unlike static evaluation approaches that primarily emphasize predictive performance, fairness reporting, or post-deployment monitoring in isolation, OADA operationalizes deployment assurance as an adaptive governance process influenced by threshold instability, disagreement amplification, subgroup-sensitive operational behaviour, remediation progression, and governance-state transitions.

The experimental evaluation demonstrated that deployment interpretation may evolve dynamically under comparatively small threshold adjustments and mitigation-sensitive operational conditions. Across both threshold-sensitive and remediation-aware evaluation settings, deployment readiness transitioned between deployable, restricted, reassessment-required, escalated-governance, and blocked-deployment states under changing operational conditions. These findings support the broader argument that operational deployment assurance cannot be reliably inferred from isolated fairness metrics or aggregate predictive performance alone.

To address these limitations, the proposed framework introduced several governance constructs, including Threshold Stability Zones (TSZ), Deployment Readiness Classification (DRC), governance-state evolution modelling, reassessment-oriented deployment progression, and remediation-aware assurance interpretation. Collectively, these mechanisms position deployment assurance as an adaptive governance process rather than a static evaluative outcome.

As AI systems continue expanding into high-stakes operational environments, governance mechanisms capable of supporting lifecycle-oriented deployment assurance, escalation-sensitive governance handling, and adaptive reassessment workflows may become increasingly necessary. OADA provides an initial operational governance foundation for supporting these deployment assurance requirements within evolving real-world AI deployment ecosystems.

The proposed framework therefore contributes toward a transition from observational AI governance toward operationally executable deployment assurance infrastructures for high-stakes AI systems.

\bibliographystyle{IEEEtran}
\bibliography{references}

@IEEEtranBSTCTL{IEEEexample:BSTcontrol,
  CTLdash_repeated_names = "no"
}

@article{b1,
  author={S. Barocas and A. D. Selbst},
  title={Big Data's Disparate Impact},
  journal={California Law Review},
  volume={104},
  number={3},
  pages={671--732},
  year={2016}
}

@inproceedings{b2,
  author={J. Buolamwini and T. Gebru},
  title={Gender Shades: Intersectional Accuracy Disparities in Commercial Gender Classification},
  booktitle={Proc. Conf. Fairness, Accountability and Transparency},
  pages={77--91},
  year={2018}
}

@inproceedings{b3,
  author={M. Hardt and E. Price and N. Srebro},
  title={Equality of Opportunity in Supervised Learning},
  booktitle={Proc. NeurIPS},
  pages={3315--3323},
  year={2016}
}

@inproceedings{b4,
  author={S. Verma and J. Rubin},
  title={Fairness Definitions Explained},
  booktitle={Proc. IEEE/ACM Int. Workshop on Software Fairness},
  pages={1--7},
  year={2018}
}

@article{b5,
  author={A. Chouldechova},
  title={Fair Prediction with Disparate Impact: A Study of Bias in Recidivism Prediction Instruments},
  journal={Big Data},
  volume={5},
  number={2},
  pages={153--163},
  year={2017}
}

@inproceedings{b6,
  author={J. Kleinberg and S. Mullainathan and M. Raghavan},
  title={Inherent Trade-Offs in the Fair Determination of Risk Scores},
  booktitle={Proc. ITCS},
  year={2017}
}

@article{b7,
  author={S. Friedler and C. Scheidegger and S. Venkatasubramanian},
  title={On the Impossibility of Fairness},
  journal={arXiv preprint arXiv:1609.07236},
  year={2016}
}

@inproceedings{b9,
  author={M. Kearns and S. Neel and A. Roth and Z. S. Wu},
  title={Preventing Fairness Gerrymandering: Auditing and Learning for Subgroup Fairness},
  booktitle={Proc. ICML},
  pages={2564--2572},
  year={2018}
}

@inproceedings{b10,
  author={M. Feldman and S. A. Friedler and J. Moeller and C. Scheidegger and S. Venkatasubramanian},
  title={Certifying and Removing Disparate Impact},
  booktitle={Proc. KDD},
  pages={259--268},
  year={2015}
}

@inproceedings{b11,
  author={B. H. Zhang and B. Lemoine and M. Mitchell},
  title={Mitigating Unwanted Biases with Adversarial Learning},
  booktitle={Proc. AIES},
  pages={335--340},
  year={2018}
}

@inproceedings{b12,
  author={G. Pleiss and M. Raghavan and F. Wu and J. Kleinberg and K. Q. Weinberger},
  title={On Fairness and Calibration},
  booktitle={Proc. NeurIPS},
  pages={5680--5689},
  year={2017}
}

@article{b13,
  author={P. J. Phillips and F. Jiang and A. Narvekar and J. Ayyad and A. J. O'Toole},
  title={An Other-Race Effect for Face Recognition Algorithms},
  journal={ACM Transactions on Applied Perception},
  volume={8},
  number={2},
  pages={1--11},
  year={2011}
}

@article{b14,
  author={I. Serna and A. Morales and J. Fierrez and N. Obradovich},
  title={Sensitive Loss: Improving Accuracy and Fairness of Face Representations with Discrimination-Aware Deep Learning},
  journal={Artificial Intelligence},
  volume={305},
  year={2022}
}

@inproceedings{b15,
  author={K. Karkkainen and J. Joo},
  title={FairFace: Face Attribute Dataset for Balanced Race, Gender, and Age},
  booktitle={Proc. WACV},
  pages={1548--1558},
  year={2021}
}

@article{b16,
  author={Z. Obermeyer and B. Powers and C. Vogeli and S. Mullainathan},
  title={Dissecting Racial Bias in an Algorithm Used to Manage the Health of Populations},
  journal={Science},
  volume={366},
  number={6464},
  pages={447--453},
  year={2019}
}

@article{b17,
  author={A. Rajkomar and J. Dean and I. Kohane},
  title={Machine Learning in Medicine},
  journal={New England Journal of Medicine},
  volume={380},
  number={14},
  pages={1347--1358},
  year={2019}
}

@techreport{b18,
  author={{National Institute of Standards and Technology}},
  title={Artificial Intelligence Risk Management Framework (AI RMF 1.0)},
  institution={NIST AI 100-1},
  year={2023}
}

@misc{b19,
  author={{European Parliament and Council of the European Union}},
  title={Regulation (EU) 2024/1689 Laying Down Harmonised Rules on Artificial Intelligence},
  year={2024}
}

@misc{b20,
  author={{ISO/IEC}},
  title={ISO/IEC 42001:2023 Artificial Intelligence --- Management System},
  year={2023}
}

@inproceedings{b21,
  author={E. Breck and others},
  title={The ML Test Score: A Rubric for ML Production Readiness and Technical Debt Reduction},
  booktitle={Proc. IEEE Big Data},
  pages={1123--1132},
  year={2017}
}

@inproceedings{b22,
  author={D. Sculley and others},
  title={Hidden Technical Debt in Machine Learning Systems},
  booktitle={Proc. NeurIPS},
  pages={2503--2511},
  year={2015}
}

@article{b23,
  author={M. Kreuzberger and N. Kuehl and S. Hirschl},
  title={Machine Learning Operations (MLOps): Overview, Definition, and Architecture},
  journal={IEEE Access},
  volume={11},
  pages={31866--31879},
  year={2023}
}

@article{b24,
  author={J. Gama and I. Zliobaite and A. Bifet and M. Pechenizkiy and A. Bouchachia},
  title={A Survey on Concept Drift Adaptation},
  journal={ACM Computing Surveys},
  volume={46},
  number={4},
  pages={1--37},
  year={2014}
}

@incollection{b25,
  author={R. Bloomfield and P. Bishop},
  title={Safety and Assurance Cases: Past, Present and Possible Future},
  booktitle={Making Systems Safer},
  pages={51--67},
  publisher={Springer},
  year={2010}
}

@inproceedings{b26,
  author={T. Kelly and R. Weaver},
  title={The Goal Structuring Notation---A Safety Argument Notation},
  booktitle={Proc. Dependable Systems and Networks Workshop},
  year={2004}
}

@misc{b27,
  author={{UK Department for Science, Innovation and Technology}},
  title={Introduction to AI Assurance},
  year={2024}
}

@article{b28,
  author={B. Mittelstadt},
  title={Principles Alone Cannot Guarantee Ethical AI},
  journal={Nature Machine Intelligence},
  volume={1},
  number={11},
  pages={501--507},
  year={2019}
}

@inproceedings{b29,
  author={S. Selbst and D. Boyd and S. Friedler and S. Venkatasubramanian and J. Vertesi},
  title={Fairness and Abstraction in Sociotechnical Systems},
  booktitle={Proc. FAT*},
  pages={59--68},
  year={2019}
}

@inproceedings{b30,
  author={M. Veale and M. Van Kleek and R. Binns},
  title={Fairness and Accountability Design Needs for Algorithmic Support in High-Stakes Public Sector Decision-Making},
  booktitle={Proc. CHI},
  year={2018}
}

@article{b31,
  author={M. Kroll and others},
  title={Accountable Algorithms},
  journal={University of Pennsylvania Law Review},
  volume={165},
  number={3},
  pages={633--705},
  year={2017}
}

@inproceedings{b32,
  author={C. Guo and G. Pleiss and Y. Sun and K. Q. Weinberger},
  title={On Calibration of Modern Neural Networks},
  booktitle={Proc. ICML},
  pages={1321--1330},
  year={2017}
}

@article{b33,
  author={A. N. Angelopoulos and S. Bates},
  title={A Gentle Introduction to Conformal Prediction and Distribution-Free Uncertainty Quantification},
  journal={Foundations and Trends in Machine Learning},
  volume={16},
  number={4},
  pages={494--591},
  year={2023}
}

@article{b34,
  author={T. Gebru and others},
  title={Datasheets for Datasets},
  journal={Communications of the ACM},
  volume={64},
  number={12},
  pages={86--92},
  year={2021}
}

@inproceedings{b35,
  author={M. Mitchell and others},
  title={Model Cards for Model Reporting},
  booktitle={Proc. FAT*},
  pages={220--229},
  year={2019}
}

@inproceedings{b36,
  author={I. Raji and others},
  title={Closing the AI Accountability Gap: Defining an End-to-End Framework for Internal Algorithmic Auditing},
  booktitle={Proc. FAccT},
  pages={33--44},
  year={2020}
}

@article{b37,
  author={A. Jobin and M. Ienca and E. Vayena},
  title={The Global Landscape of AI Ethics Guidelines},
  journal={Nature Machine Intelligence},
  volume={1},
  pages={389--399},
  year={2019}
}

@article{b38,
  author={Kkhalid A. Alsayed},
  title={Why Aggregate Accuracy Is Inadequate for Evaluating Fairness in Law Enforcement Facial Recognition Systems},
  journal={arXiv preprint arXiv:2603.28675},
  year={2026}
}

@article{b39,
  author={Khalid Adnan Alsayed},
  title={When Fairness Metrics Disagree: Evaluating the Reliability of Demographic Fairness Assessment in Machine Learning},
  journal={arXiv preprint arXiv:2604.15038},
  year={2026}
}

@misc{b40,
  author={K. A. Alsayed},
  title={When AI Gets It Wrong: Reliability and Risk in AI Assisted Medication Decision Systems},
  howpublished={arXiv preprint arXiv:2604.01449},
  year={2026}
}
\end{document}